\documentclass[11pt]{article}

\usepackage[final]{acl}

\usepackage{times}
\usepackage{latexsym}

\usepackage[T2A,T1]{fontenc}
\usepackage[utf8]{inputenc}

\usepackage{microtype}

\usepackage{inconsolata}

\usepackage{graphicx}
\usepackage{xspace}
\usepackage{enumitem}
\usepackage{booktabs}
\usepackage{tabularx}
\usepackage{multirow}
\usepackage{makecell}
\usepackage[serbian,english]{babel}
\usepackage{threeparttable}
\usepackage{dblfloatfix}

\usepackage{todonotes}
\makeatletter
\newcommand*\iftodonotes{\if@todonotes@disabled\expandafter\@secondoftwo\else\expandafter\@firstoftwo\fi}
\makeatother

\newcommand{\eclektic}{ECLeKTic\xspace}
\newcommand{\multiloko}{MultiLoKo\xspace}
\newcommand{\belebele}{Belebele\xspace}
\newcommand{\gpt}{GPT-OSS-120B\xspace}
\newcommand{\glm}{GLM-4.5-Air\xspace}
\newcommand{\qwenL}{Qwen3-235B-A22B\xspace}
\newcommand{\qwenM}{Qwen3-30B-A3B\xspace}
\newcommand{\olmo}{Olmo-3.1-32B\xspace}
\newcommand{\gemini}{Gemini\xspace}
\title{Large Reasoning Models Struggle to Transfer Parametric Knowledge Across Scripts}

\author{Lucas Bandarkar\textsuperscript{$\mathsection$} \ \ \ \ \ \ \ \ \ \ \ \ Alan Ansell \ \ \ \ \ \ \ \ \ \ \ \ Trevor Cohn\textsuperscript{$\phi$} \\
        Google Research \\
        \textsuperscript{$\mathsection$}University of California, Los Angeles \ \ \ \ \textsuperscript{$\phi$}University of Melbourne}

\begin{document}
\maketitle
\begin{abstract}
In this work, we analyze shortcomings in cross-lingual knowledge transfer in large, modern reasoning LLMs. We demonstrate that the perceived gap in knowledge transfer is primarily a script barrier.
First, we conduct an observational data analysis on the performance of thinking models on two datasets with local knowledge from around the world, \eclektic and \multiloko.
Our regression analysis shows that script match—not language or family—is the primary predictor of knowledge transfer failure once model capability and question difficulty are accounted for.
We further this finding by providing the LLMs with the key entities of the questions in their source language
and find that this disproportionately improves cross-script questions. We then posit that these LLMs could be reasoning better at test-time. To evaluate this, we develop a synthetic generation pipeline to design SFT samples to encourage the model to better reason about transliteration ambiguities when trying to fetch parametric knowledge at inference-time. We show that teaching two models to reason better reduces the cross-script transfer gap. As a result, we conclude that there is potential to improve cross-lingual parametric knowledge transfer during post-training.
\end{abstract}

\section{Introduction}

The recent scaling of parameters and training data has increased the multilingual capabilities of Large Language Models (LLMs), leading to remarkable cross-lingual transfer across domains and tasks \citep{hu-etal-2025-quantifying, he-etal-2025-scaling}. While these modern models exhibit a huge capacity to learn, retain, and retrieve factual global knowledge, issues regarding inconsistency across languages remain \citep{romanou2025include, wang-etal-2025-lost-multilinguality,guo2025liveclktbenchreliableevaluationcrosslingual}.
% This work identifies that, with respect to knowledge, language script---specifically the mismatch between the script of the query and the script of origin of the knowledge---remains a notable barrier.
% Specifically, that is, the mismatch between the script of the query and the script of origin of the knowledge (e.g. a query in English about a Serbian monastery).

In this work, we analyze the shortcomings in cross-lingual knowledge transfer within large, modern reasoning LLMs and find that language script remains a notable barrier. Inconsistency is higher when there is a mismatch between the script of the query and the script of origin of the knowledge (e.g. a query in English about a Serbian monastery). Utilizing two local knowledge QA evaluation datasets, \eclektic \citep{eclektic} and \multiloko \citep{hupkes2025multiloko}, we first conduct an observational data analysis across numerous thinking models in Section~\ref{obs}. We find that when controlling for the statistical difficulty of questions and the model's inherent capability (using reading comprehension results as a proxy), cross-lingual gaps are statistically insignificant when the source and target languages share a script. However, languages not sharing a script leads to a 13\% lower accuracy, implying significant transfer limitations when retrieving knowledge across scripts. 
In Section~\ref{prefix}, we identify cross-lingual mapping of entities as one of the drivers of this script barrier. We show that prefixing a model's English thinking trace with the original-language entity disproportionately improves performance on cross-script questions, highlighting a specific failure in cross-script entity alignment.

Finally, positing that reasoning LLMs possess the latent capacity to better reason across scripts, we develop a synthetic generation pipeline to design Supervised Fine-Tuning (SFT) samples. We demonstrate that teaching the model to explicitly reason about transliteration ambiguities during inference significantly reduces the cross-script transfer gap. This suggests a large proportion of the gap from cross-script limitations can cheaply be solved.

\section{Related Work}

To address constraints to model vocabulary, numerous works proposed different uses of transliteration as test-time solutions for cross-script transfer limitations \citep{amrhein-sennrich-2020-romanization, muller-etal-2021-unseen, purkayastha-etal-2023-romanization, j-etal-2024-romansetu}.
More recent works on LLMs identify persistent cross-lingual inconsistency, especially in knowledge \citep{wang-etal-2025-lost-multilinguality}, with some noting script's high influence \citep{beniwal-etal-2024-cross, nie-etal-2025-bmike}. Solutions often involve better aligning representations to enhance cross-lingual sharing and transfer \citep{lu-etal-2025-paths, sun_aaai}. In analyses of knowledge representations, both \citet{saji-etal-2025-romanlens} and \citet{ifergan-etal-2025-beneath} identify script-specific representations.

% \section{Simple Model for Cross-Lingual Knowledge}

% In order to correctly answer knowledge-intensive multilingual queries, an LLM must accomplish numerous steps. While the underlying computations are inherently complex, we propose a simplified sequential model of these required steps, expanding upon the decomposition from \citet{hu-etal-2025-large-language}:
% \begin{enumerate}[nosep, leftmargin=*]
%     \item The knowledge required is present somewhere in model parameters.
%     \item The LLM understands the multilingual question.
%     \item The LLM accesses the parametric knowledge. 
%     \item The LLM reasons properly over the question and fetched knowledge.
%     \item The LLM samples the correct answer in the requested format and language during generation.
% \end{enumerate}

%% needs a re-write
\section{Setup}
We work with five state-of-the-art reasoning LLMs, \qwenM\footnote{Specifically, \qwenM-Thinking-2507}, \qwenL\footnote{Specifically, \qwenL-Thinking-2507} \citep{qwen3}, \glm \citep{glm}, \gpt \citep{gptoss}, and \olmo\footnote{Specifically, \olmo-Think} \citep{olmo}. While the Qwen3 series is the most multilingual, the 30-32B models naturally lag the larger models in performance. 
% All are sparse mixture-of-experts except \olmo. % is the slowest at inference-time because of its dense architecture.

We focus on two cross-lingual knowledge QA datasets, \multiloko \citep{hupkes2025multiloko} and \eclektic \citep{eclektic}. While designed differently, both datasets attempt to ask questions about rare knowledge likely only present in one or very few languages in an LLMs' training corpus. Each question has one labeled source language, which allows us to study the cross-lingual transfer dynamics in directions beyond Eng->X. We use \emph{cross-script questions} to refer to questions where the language of the question differs from the source language.
We note that across many runs with slightly perturbed conditions (prompts, hyperparameters, etc.), there was significantly more variance for individual questions in low-resource languages, in line with \citet{piratla2025rethinkingcrosslingualgapsstatistical}.

% delete if no space
% Our exploratory analysis reveals that shorter thinking traces correlate strongly with higher accuracy, indicating that in all languages, excessive thinking reflects model confusion \citep{wu2025when}.

\section{Observational Data Analysis} \label{obs}

To determine which factors most limit a model's ability to use its knowledge across different languages, we conduct a regression analysis on \eclektic and \multiloko with our table of model results for each question. For features, we start with indicators for the model and the source language to account for different model capabilities and question subset difficulties.
Rather than relying on categorical indicators for the language of the question (i.e., target language), we use the model's accuracy ($\in$ [0,1]) on the Belebele reading comprehension task \citep{bandarkar-etal-2024-belebele} as a continuous feature. This serves as a direct proxy for the model's "raw" linguistic capability in that language\footnote{The task of reading comprehension requires no external knowledge or reasoning, since the context is provided. As a result, it is solely dependent on textual understanding.}, reducing the feature space and enhancing interpretability. We argue this captures cross-lingual proficiency more granularly than language-specific fixed effects all while increasing degrees of freedom for our regression model.
% Rather than an indicator for the language of the question (i.e., target language), we have the model's accuracy $\in$ [0,1] in reading comprehension in that language (using \belebele \citep{bandarkar-etal-2024-belebele}). We use this as a proxy for model's pure linguistic capabilities, replacing many potential indicators into one feature.
And finally, we add binary variables to represent the relationship between the source and target languages: whether they are the same language family\footnote{Using the classification in FLORES \citep{flores101}}, script, or identity. Note that since we do not use text-level features, many questions shared the same set of features. For \eclektic, this resulted in 144 unique set of features with an average of 32 questions each, while for \multiloko, this resulted in 91 unique sets with exactly 250 questions each.

\begin{table}[h]
    \centering
    % \small
    \setlength{\tabcolsep}{0pt}
    \begin{threeparttable}

    \caption{OLS Coefficients ($\beta$) when trying to predict the binary score of a question, \textbf{with no scaling}. \textit{n.s.} denotes statistically insignificant ($\beta$); $p > 0.05$.}
    \begin{tabular}{@{}lcc@{}}
    \toprule
    \textbf{Variable} & \textbf{ECleKTic} & \textbf{MultiLoKo} \\ 
    OLS Intercept                  & \normalsize{$-$0.350}            & \normalsize{$-$0.439}   \\
    \midrule
    Belebele Score                 & \normalsize{0.999}            & \normalsize{0.937}            \\
    \midrule
    Script Match                   & \normalsize{0.121}            & \normalsize{0.133}            \\
    Language Family                & \normalsize{\textit{n.s.}}    & \normalsize{\textit{n.s.}}    \\
    In-Language                    & \normalsize{\textit{n.s.}}    & \normalsize{\textit{n.s.}}    \\
    \midrule
    \multicolumn{3}{l}{\textit{Model Indicators (vs. Qwen3-30B)}}                 \\
    \quad GPT-OSS-120B             & \normalsize{0.119}           & \normalsize{0.065}           \\
    \quad Olmo-3.1-32B             & \normalsize{0.061}           & \normalsize{0.091}           \\
    \quad GLM-4.5-Air              & \normalsize{0.040}           & \normalsize{\textit{n.s.}}    \\
    \quad Qwen3-235B               & \normalsize{0.038}          & \normalsize{0.031}           \\
    \midrule
%     \textit{Src. Lang (vs. EN)}             & In Range:        & In Range: \\
%                                 & \small{[-0.300, -0.059]}  & \small{[\textit{n.s.}, 0.330]} \\
%     \bottomrule
%     \end{tabular}

    \multicolumn{3}{l}{\textit{Src Lang Indicators\tnote{$\dagger$}} \ (vs. EN)} \\
    & \small{[-0.300, -0.059]}  & \small{[\textit{n.s.}, 0.330]} \\
    \bottomrule
    \end{tabular}
    \begin{tablenotes}
        \item[$\dagger$] \small Range of $\beta$ for source-language subset indicators used to account for subset difficulty.
    \end{tablenotes}
    \label{coef_unscaled}
    \end{threeparttable}

\end{table}

Using these features, we fit an Ordinary Least Squares (OLS) regression model to predict the binary feature of the actual model score (1.0 or 0.0) on a given question given the features to disentangle the drivers of cross-lingual gaps (See full details in Appendix~\ref{ols}, including a discussion on OLS vs. logistic regression). Although we fit separate models for \eclektic and \multiloko, the results are remarkably consistent as displayed in Table~\ref{coef_unscaled}. The \belebele score—our proxy for "raw" linguistic capability—has the highest coefficient but also is the only continuous variable (with a much smaller standard deviation), meaning it is not directly comparable to the indicator variables. However, in Table~\ref{coef} we provide coefficients with all features standard-scaled, and show that \belebele~score captures the most variance in general model capabilities. The fact that simple linguistic understanding is the dominant predictor supports the continued utility of translation evaluations as proxies for multilingual task benchmarks \citep{issaka2026translationscalableproxymultilingual}. 

Most notably, however, whether or not the source and target language were the same, or even in the same family, has a statistically \textit{in}significant coefficient $\beta$\footnote{Concretely, the null hypothesis $\beta = 0$ cannot be rejected.} when regressed alongside script match. This indicator has a very high $\beta$, implying a 12.1\% and 13.3\% increase in probability of correct response if the target and source languages share a script. This demonstrates significant transfer limitations in retrieving knowledge across scripts and that, when controlling for this, transfer across languages within the same script, or even family is only limited by the model's ``raw'' performance in the target language. For now, this is purely observational and so we explore this further.

\section{Prefixing Thinking} \label{prefix}

We evaluate the hypothesis that cross-script gaps stem in part from a model’s inability to correctly link key entities across languages. We do so by pre-filling the beginning of each model's thinking trace and providing the key entity
% \alan{you have gone from ``key entities'' to ``THE specific entity''. please clarify}
in the source language. This eliminates the potential source of retrieval error from failed cross-lingual entity mappings.

We start by using \gemini-3-Pro \citep{gemini3} to identify important entities (notably named entities) in all questions in their original source language. We prompt each model and force its reasoning to begin with a specific prefix similar to "<think> Okay, the key entity of the question in its source language is \{src\_entity\}". As a result, the model is provided the pre-translation entities central to the question. 
Note that, irrespective of the language of the question, all 5 models almost always produce English thinking tokens. GLM and Qwen3 models, however, sometimes produce Chinese thinking when prompted in Chinese. In this experiment, we implicitly force it to think in English. As a control,
% \alan{is control the right word for this?}, % lucas: i guess that's what I've been using it as, open to rewording suggestions
we also evaluate the model with that entire prefix minus the actual "\{src\_entity\}" at the end. To maintain the natural distribution, we stylize the prefix in accordance with the wording typically used by each model to start its thinking (provided in Appendix~\ref{prefixes}). By providing the entity as an anchor, we test whether the primary bottleneck is the cross-lingual entity mapping.

As expected, providing the pre-translation entity of the question resulted in positive $\Delta$ on \eclektic and \multiloko accuracy. On all types of questions, we notice a small uptick on average, indicating that cross-lingual mapping of the entities was a limitation for such cross-lingual knowledge retrieval, as displayed in Table~\ref{prefixtbl}. 

\begin{table}[h]
\centering
% \small
\setlength{\tabcolsep}{2pt} % Shrinks padding between columns to save space
\begin{tabularx}{\columnwidth}{@{} X c|cc|cc @{}}
\toprule
\textbf{Condition} & \textbf{All} & \makecell{\textbf{Non-}\\\textbf{Latin}} & \makecell{\textbf{Latin}} & \makecell{\textbf{Script}\\\textbf{$\neq$}} & \makecell{\textbf{Script}\\\textbf{=}} \\
\midrule
\multicolumn{6}{l}{\textit{\gpt}} \\
Original                & 20.6 & 16.3 & 23.6 & 18.5 & 24.6 \\
Prefix, Control         & 21.2 & 17.9 & 23.5 & 19.3 & 24.8 \\
Prefix w/ src Ent.   & 22.2 & 19.4 & 24.2 & 20.7 & 25.1 \\
\hspace{2em} $\Delta$   & +1.0 & +1.5 & +0.7 & +1.4 & +0.2 \\
\midrule
\multicolumn{6}{l}{\textit{\qwenL}} \\
Original                & 28.3 & 24.2 & 31.2 & 26.0 & 32.8 \\
Prefix, Control         & 28.3 & 24.5 & 31.1 & 26.1 & 32.7 \\
Prefix w/ src Ent.   & 30.0 & 27.1 & 32.2 & 28.6 & 32.7 \\
\hspace{2em} $\Delta$   & +1.7 & +2.6 & +1.1 & +2.5 & +0.0 \\
\midrule
\multicolumn{6}{l}{\textit{\glm}} \\
Original                & 25.5 & 21.2 & 28.6 & 22.6 & 29.6 \\
Prefix, Control         & 25.2 & 20.5 & 28.6 & 22.6 & 30.1 \\
Prefix w/ src Ent.   & 26.9 & 22.8 & 29.9 & 24.7 & 31.2 \\
\hspace{2em} $\Delta$   & +1.7 & +2.3 & +1.3& +2.1 & +1.1 \\
\midrule
Average $\Delta$        & +1.5 & +2.1 & +1.0 & +2.1 & +0.5\\
\bottomrule
\end{tabularx}
\caption{Impact of our think-prefixing on \eclektic accuracy $\uparrow$. Script $\neq$ denotes cross-script; Script $=$ denotes same-script.}
\label{prefixtbl}
\end{table}
\begin{table*}[b]
\centering
\small
\setlength{\tabcolsep}{2pt} % Tighten column padding to fit 18 columns
\begin{tabular*}{\textwidth}{@{\extracolsep{\fill}} l l | c | cc | cc | cc | c | cc | cc | cc @{}}
\toprule
& & \multicolumn{7}{c|}{\textbf{\eclektic} (12 languages)} & \multicolumn{7}{c}{\textbf{\multiloko} (31 languages)} \\
\cmidrule{3-9} \cmidrule{10-16}
& & & \multicolumn{2}{c|}{Latin} & \multicolumn{2}{c|}{Resource} & \multicolumn{2}{c|}{Script} & & \multicolumn{2}{c|}{Latin} & \multicolumn{2}{c|}{Resource} & \multicolumn{2}{c}{Script} \\
\textbf{Model} & \textbf{Run} & \textbf{All} & \textbf{No} & \textbf{Latin} & \textbf{Low} & \textbf{High} & \textbf{$\neq$} & \textbf{=} & \textbf{All} & \textbf{No} & \textbf{Latin} & \textbf{Low} & \textbf{High} & \textbf{$\neq$} & \textbf{=} \\
\midrule

% --- GPT ---
\multirow{2}{*}{\gpt} 
 & Base & 20.6 & 16.3 & 23.6 & 12.6 &  23.9 & 18.5 & 24.6 & 
 24.4& 18.0 & 31.2 & 17.0 & 27.8 & 21.6 & 27.3  \\
 & LoRA  & 21.6 & 18.6 & 23.8 & 15.3 & 24.0 & 20.1 & 24.7 & 
 25.0 & 18.7 & 32.0 & 17.7 & 28.3 & 22.3 & 27.4 \\
\midrule

% --- Qwen ---
\multirow{2}{*}{\qwenM} 
 & Base & 18.1 & 15.1 & 20.1 & 12.2 & 20.0 & 16.9 & 20.3 & 
 25.6 & 21.4 & 30.1 & 19.6 & 27.1 & 21.1 & 28.0 \\
 & LoRA  & 18.9 & 16.3 & 20.8 & 12.8 & 18.2 & 18.0 & 20.6 & 
 26.0 & 21.9 & 30.4 & 20.2 & 27.4 & 21.7 & 28.2 \\
% \midrule

% % --- GLM ---
% \multirow{2}{*}{\olmo} 
%  & Base & 0.0 & 0.0 & 0.0 & 0.0 & 0.0 & 0.0 & 0.0 & 0.0 & 0.0 & 0.0 & 0.0 & 0.0 & 0.0 & 0.0 \\
%  & LoRA  & 0.0 & 0.0 & 0.0 & 0.0 & 0.0 & 0.0 & 0.0 & 0.0 & 0.0 & 0.0 & 0.0 & 0.0 & 0.0 & 0.0 \\

\bottomrule
\end{tabular*}
\caption{Comprehensive evaluation of Base vs. SFT across two datasets. We compare performance across Latinity (Non-Latin vs. Latin), Resource levels (Low vs. High), and Script matching ($\neq$ vs. $=$).}
\label{sft_results}
\end{table*}

However, this sort of prefixing seems to disproportionately impact cross-script questions. For example, for \qwenL, the impact from prefixing the pre-translation entity is negligible for all other questions, while there is a 2.5-point boost for cross-script questions. Averaged across three models, the $\Delta$ for cross-script questions is $+2.1$ while being just $+0.5$ for the rest. This further confirms our hypothesis that script is a barrier for these models, negatively impacting knowledge transfer. 
% See Appendix~\ref{prefixtbl} for detailed results. % are provided in Table~\ref{prefixtbl} in the appendix.

\section{Reducing Gaps using SFT} \label{sft_experiments}

Based on qualitative and quantitative analyses of thinking traces in Sections~\ref{obs} and \ref{prefix}, we hypothesize that these LLMs have the capacity to better reason over the potential transliteration ambiguities in the question.
% added for arxiv version:
For example, we see in Table~\ref{prefixtbl} that the prefix \textit{without} the source entity provided (labeled as "Prefix, Control", see Appendix~\ref{prefixes}) often leads to some increase in performance for \gpt. This simple modification to the beginning of the thinking trace implies there is a way to generally improve these model's reasoning.

To evaluate this, we experiment with teaching the model to reflect carefully about what entity the translated question could be referring to. This is done via LoRA Supervised-Finetuning (SFT) on two of the smaller (but still large) models \qwenM and \gpt.

\subsection{Synthetic Data Generation}

To create SFT data, we use a multi-step pipeline that uses \gemini to generate the reasoning logic that we want to teach the models. However, due to the sensitivity of SFT to formatting \citep{zhou2023lima, chu2025sft, matsutani2026rl}, we additionally use style transfer to ensure the training samples resemble each model's own style and formatting. This stylistic alignment ensures the model can learn the reasoning patterns without the distraction of unfamiliar formatting.

% removed nosep for arxiv
\begin{enumerate}[leftmargin=*,]
    \item We prompt \gemini to create multilingual knowledge questions and answers. Alongside, we instruct \gemini to create a thinking trace where the reasoner reflects carefully about the potential transformations during translation that could have led to the resulting entity in the question. We iterate through prompts and filtering mechanisms until qualitative checks of the generated questions reveal quality samples. In the end, we use a variety of prompts to induce more variation (an example is provided in Appendix~\ref{geminiprompts}).
    \item Using only the question generated by \gemini, we prompt each of the three models to answer the questions, without any special instructions.
    \item Then, we provide the \gemini question, thinking, answer in context and prompt the target model to regenerate the thinking trace in its own style. We provide its own unadulterated thinking trace on that question to demonstrate the difference in style between it and \gemini. We provide the exact instruction in Appendix~\ref{styletransferprompts}.
\end{enumerate}

\noindent We provide additional details on this data generation pipeline in Appendix~\ref{datagendetails}. The resulting data is in a variety of languages, but as specified in our instructions, the question and answer are the same language and the thinking is English.

\subsection{SFT Results}
Then, we perform LoRA SFT using Llama-Factory \citep{llamafactory} using these training samples. We provide specifics in Appendix~\ref{loradetails}. Our qualitative analysis of the thinking traces reveals a change in the first few sentences of the thinking trace to reflect over possible transliterations, as desired. % EXPAND HERE 

As displayed in Table~\ref{sft_results}, we can achieve modest improvement across both models, evidenced on both \eclektic and \multiloko. Through a simple SFT of just a few hundred samples and 8 epochs, we are able to teach the model to better stop and think about the potential entity of origin of the question. While this does not help the model gain more knowledge, the fact that simply reasoning differently enables a few point boost shows that its cross-lingual reasoning behavior is not optimal. Notably, the performance increase happens significantly more with low-resource languages and on cross-script questions, as was hypothesized. This suggests that these model's barrier around script is not necessarily a fundamental one, but one relating to its internal retrieval mechanisms.
\section{Discussion and Conclusion}

Our investigation into the cross-lingual capabilities of state-of-the-art reasoning LLMs reveals that the ``language barrier'' in knowledge retrieval is often, more accurately, a ``script barrier'', most notably defined in Section~\ref{obs}. Intuitively, this could be expected. When a query and the relevant parametric knowledge share a script (e.g., Spanish to Indonesian), there is likely subword token overlap that allow for easy recall. Conversely, in cross-script scenarios, disjoint token sets force the model to rely entirely on abstract, semantic mappings. Even so, the persistence of this barrier in massive reasoning models is surprising. Current literature suggests that as models scale, they develop language-agnostic representations, parameterizing knowledge redundantly in a way that should be robust to surface-level shifts \citep{lim2025languagespecificlatentprocesshinders, chen2025the, dumas-etal-2025-separating, bandarkar2026multilingual}. In addition, these models are all generating English thinking traces. Even so, our results suggest that knowledge tasks remain subject to syntax-induced latent processes \citep{hu-etal-2025-large-language}, as our think prefixing experiments in Section~\ref{prefix} show cross-script entity mappings are a cause for this gap.

Crucially, our experiments reveal that this struggle is not merely one of representation, but of reasoning. The success of our simple SFT runs in Section~\ref{sft_experiments} demonstrates that LLMs sometimes possess the requisite parametric knowledge but fail to access it during reasoning. By simply teaching the model to reflect more carefully about transliteration ambiguities and entity origins, we achieved modest improvements across models. 

Consequently, our results suggest that the most accessible solution for reducing cross-lingual knowledge gaps lies in post-training reasoning improvements. While we utilized SFT due to resource constraints, we hypothesize that Reinforcement Learning (RL) could more effectively elicit these specific reasoning behaviors. However, relying solely on inference-time reasoning is likely insufficient; fundamental improvements must also come from ensuring knowledge representations are more distant from token-level forms, an area requiring further research. Ultimately, we identify that the largest potential for gain lies at the intersection of low-resource languages and cross-script settings, where simple reasoning interventions can unlock significant latent knowledge.

\section*{Limitations}

\paragraph{Real-World Transliteration Ambiguities} A factor potentially weighing on cross-script issues is from the evaluation dataset construction. In the translation of questions, it is possible that translation ambiguity and even errors are higher for cross-script translations. This could arise from rare entities from one part of the world not having transliterations that are well-agreed upon. However, this would affect our \textit{observational} regression analysis in Section~\ref{obs}. Our next two controlled experiments analyses in Sections~\ref{prefix} and~\ref{sft_experiments}, analyze the $\Delta$ and find it to be higher to cross-script questions, which is separate from inherent question difficulty due to transliteration ambiguities.

\paragraph{SFT rather than RL} For Section~\ref{sft_experiments}, we acknowledge that Reinforcement Learning methods are currently the state-of-the-art for eliciting particular reasoning behavior, but we elect LoRA Supervised-Finetuning (SFT) due to resource constraints.

%%% REMOVE FOR ARXIV VERSION
% \section*{Use of AI Assistants}

% We used Gemini for assistance with drafting language in the paper, code development, and research ideation. Any output used was reviewed and edited by the authors, who remain fully responsible for the content.

%%% COMMENT OUT FOR REVIEW !
\section*{Acknowledgments}

We acknowledge Vihari Piratla, Eleftheria Briakou, Guido Zuccon, and Raj Dabre for their feedback and support during this work.

\bibliography{custom}

\appendix
% \section{Data Generation Details} \label{datagen}
\section{Evaluation Setup Details} \label{eval_details}

\eclektic and \multiloko are evaluated using an auto-rater setup, using vLLM \citep{vllm} for inference. We prompt the questions with the system instruction "In just a few words, answer in the language of the question." and each of the model's individuals chat formats. \gpt is set to "Reasoning: medium". Under certain conditions, such as the model over-thinking before the max\_model\_len is exhausted, we reformat the thinking trace and prompt the model to just generate the answer. Then, the model output (and not its thinking trace) is provided to an auto-rater (implemented with \qwenM) along with the golden answer. Our auto-rater system instruction requires a one-word answer ("YES" or "NO") and provides context about the question and answer.  Our auto-rater system instructions are thoroughly tuned via qualitative analysis to ensure reliability.

\belebele (used in Section~\ref{obs}) is evaluated using the lm-evaluation-harness \citep{eval-harness} in python using the vLLM backend.

\section{OLS Details} \label{ols}

\paragraph{OLS vs. Logistic Regression}
Despite the predictor variable being a binary outcome (1.0 for correct answer, 0.0 for incorrect), we elect OLS for interpretability of the coefficients. In OLS, the coefficients represent direct probability changes rather than log-odds ratios. This approach is standard when predicted probabilities remain well within [0,1] \citep{Angrist2009}, which they do for all rows in the table. We have additional ran a logistic regression model and calculated average marginal effects. The average marginal effects (AME) remains very similar for the script match indicator: 0.124 for \eclektic an 0.137 for \multiloko, in comparison to the OLS $\beta$ of 0.121 and 0.133 reported. Importantly, the coefficients for language match and family match are also statistically insignificant (cannot reject null hypothesis that they are zero).

\paragraph{Implementation} We implement our regression models in both statsmodels \citep{seabold2010statsmodels} and scikit-learn \citep{scikit-learn}, which provides us a sanity check that the fitted model is unambiguous and deterministic. The final predictor variable and all the indicator features are binary variables with values 1.0 and 0.0. The \belebele score is the only continuous variable, and most values are in the range 0.50 to 0.90.

\paragraph{Feature Collinearity} While the features were certainly not independent and showed expected inter-correlations (e.g., between the cross-script and cross-language indicator), we verified the stability of our estimates using Variance Inflation Factors (VIF), ensuring that multicollinearity did not undermine the reliability of the observed coefficients. However, this leads to counter-intuitive coefficients such as \qwenL having the lowest $\beta$ other than \qwenM despite it being the best model by quite a bit on both datasets. This is explained by the correlation between model capability and \belebele score, which likely captures most of the variance in model capabilities.

% \begin{table}[ht]
%     \centering
%     \small
%     \caption{OLS Regression Results. Coefficients represent the impact on subset-level accuracy. Model effects are relative to Qwen3-30B-A3B. \textit{n.s.} indicates $p > 0.05$. Note that Qwen3-235B's raw strength is primarily captured by the Belebele proxy.}
%     \label{tab:ols_results}
%     \begin{tabular}{@{}lcc@{}}
%     \toprule
%     \textbf{Variable} & \textbf{ECleKTic ($\beta$)} & \textbf{MultiLoKo ($\beta$)} \\ 
%     \textbf{(Intercept)}           & \textbf{0.452} & \textbf{0.569}   \\
%     \midrule
%     Belebele Score                 & 0.088           & 0.100            \\
%     \midrule
%     Script Match                   & 0.056           & 0.064            \\
%     Language Family                & \textit{n.s.}   & \textit{n.s.}    \\
%     In-Language                    & \textit{n.s.}   & \textit{n.s.}    \\
%     \midrule
%     \textit{Model (vs. Qwen3-30B)} &                 &                  \\
%     \quad GPT-OSS-120B             & +0.046          & +0.026           \\
%     \quad Olmo-3.1-32B             & +0.022          & +0.034           \\
%     \quad GLM-4.5-Air              & +0.016          & \textit{n.s.}    \\
%     \quad Qwen3-235B               & +0.016          & +0.013           \\
%     \midrule
%     Source Language Range          & [$-0.06, -0.02$] & [$+0.01, +0.05$] \\
%     \bottomrule
%     \end{tabular}
% \end{table}

\begin{table}[h]
    \centering
    % \small
    \setlength{\tabcolsep}{0pt}
    \begin{threeparttable}

    \caption{OLS Coefficients ($\beta$) when trying to predict the binary score of a question, \textbf{after features are standard-scaled}. \textit{n.s.} denotes statistically insignificant coefficients; $p > 0.05$. The values represent the expected increase in probability from an increase in one standard deviation of the feature. It allows for comparison between continuous features (e.g. \belebele score) and the rest. The value is essentially a metric for how much variance the feature encapsulates.}
    \begin{tabular}{@{}lcc@{}}
    \toprule
    \textbf{Variable} & \textbf{ECleKTic} & \textbf{MultiLoKo} \\ 
    OLS Intercept                    & 0.452            & 0.569   \\
    \midrule
    Belebele Score                 & 0.088            & 0.100            \\
    \midrule
    Script Match                   & 0.056            & 0.064            \\
    Language Family                & \textit{n.s.}    & \textit{n.s.}    \\
    In-Language                    & \textit{n.s.}    & \textit{n.s.}    \\
    \midrule
    \multicolumn{3}{l}{\textit{Model Indicators (vs. Qwen3-30B)}}         \\
    \quad GPT-OSS-120B             & 0.046           & 0.026           \\
    \quad Olmo-3.1-32B             & 0.022           & 0.034           \\
    \quad GLM-4.5-Air              & 0.016           & \textit{n.s.}    \\
    \quad Qwen3-235B               & 0.016           & 0.013           \\
    \midrule
    % \textit{Src. Lang (vs. EN)}             & In Range:        & In Range: \\
    %                             & \small{[$-$0.064, $-$0.023]}  & \small{[0.007, 0.048]} \\
    % \bottomrule
    \multicolumn{3}{l}{\textit{Src Lang Indicators\tnote{$\dagger$}} \ (vs. EN)} \\
    & \small{[$-$0.064, $-$0.023]}  & \small{[0.007, 0.048]} \\
    \bottomrule
    \end{tabular}
    \begin{tablenotes}
        \item[$\dagger$] \small Range of $\beta$ for source-language subset indicators used to account for subset difficulty.
    \end{tablenotes}
    \label{coef}
    \end{threeparttable}
\end{table}

Interestingly, in \eclektic, the English-source subset seemed to be the easiest (the coefficient for all other languages is negative), while for \multiloko the English-source subset was the most difficult (coefficient for all other languages is positive).

% TO FINISH

\section{Think Prefixing Prompts} \label{prefixes}

Note that in certain questions in \eclektic and \multiloko, there may be more than one entity. So we create a prompt for both scenarios:

For Qwen3 and \olmo:
\begin{quote}
    "Okay, the key entity of the question in its source language is "
    
    "Okay, the important entities in the question in their source language are "
\end{quote}

For \glm:
\begin{quote}
    "The question is asking about the entity (in English) "
    
    "The question is asking about the entities (in English) "
\end{quote}

For \gpt, where \{lang\_tgt\} is the English name for the language of the question:
\begin{quote}
    "The user asks in \{lang\_tgt\} about an entity, that in its source language would be "

    "The user asks in \{lang\_tgt\} a question in multiple entites, that would be, in their source language,  "
\end{quote}

% \section{Think Prefixing Results} \label{prefixtbl}
% \input{tables/prefix_results}

% This table reveals that the increase in accuracy from providing the original source-language entities improves cross-script questions substantially more than same-script questions. In addition, we note that the incomplete prefix (labeled as "Prefix, Control", See Appendix~\ref{prefixes}) often leads to some increase in performance. This leads us to our hypothesis that the models could reason better at inference-time.

\section{Additional SFT Data Generation Details} \label{datagendetails}

We provide here details for our SFT data generation pipeline discussed in Section~\ref{sft_experiments}. We first use the prompts described in Appendix~\ref{geminiprompts} to create candidate question/thinking/answer trios with \gemini. While a quick check of some samples revealed that Gemini was creating valid answers for the questions. However, we do not particularly care if the answer is right, as our goal is not to teach the model facts. The goal is to teach it a reasoning style. We then prompt the target model with just the question, and again, do not care if its answer is coherent with \gemini's provided answer as all we care about here is the thinking \textit{style}.

We then use the prompt detailed in Appendix~\ref{styletransferprompts} to get the target model to provide a style-adhering version of the thinking trace. We are forced to discard many samples during each step if there are errors in the formatting (i.e., requested JSON formatting). Because of the higher amount of samples dropped for \qwenM during the style transfer process, we end up with 599 training samples for \gpt and only 387 training samples for \qwenM.

\section{Sample \gemini Prompt} \label{geminiprompts}

Below we provide a sample prompt used for generating question/thinking/answer trios. We check that Gemini is generating a thinking trace that does as we ask. To settle on the final prompt, it took numerous attempts to get the reasoning logic that we desire. Even after settling on this prompt, we do continuous sanity checks that the generation is coherent and logical. We evaluated a few dozen before scaling the generation to about 700 thinking traces. To induce more variability, we modify the wording of the following prompt, but the details and instructions remain the same.

\begin{quote}
I have this cross-lingual knowledge database with questions and want to fine-tune the model to reason better. The questions are often about entities and they exist in a bunch of languages in parallel. However the (human) translation of the entities is sometimes a bit ambiguous because they are very rare/obscure entities and may not have a proper written format in other scripts. Given this, can you produce seven sample model thinking traces where the model reasons about the entity and what the original entity could possibly be (potentially from different languages/scripts). Then once the model thinks it has found the entity, it provides its answers to the question in the original language ?

Please generate nothing but a JSON, where each item in the list is a dictionary containing question, thinking, and model\_response. Please don't include formatted markdown text in the thinking trace, simply a few paragraphs.
\end{quote}

\section{Style Transfer Prompt} \label{styletransferprompts}

The prompt provided to \qwenM for ``style transfer'' from Gemini's thinking trace to its style. The prompt for \gpt was worded differently because it was not following these instructions well.  \{gemini\_thinking\_trace\} and \{style\_reference\} are placeholders for the question-specific thinking trace provided by \gemini and the target model, respectively.
\begin{quote}
I am performing style transfer of a thinking trace to create SFT data for a thinking model.

<instruction>

1. Read the logic in <logic\_to\_rewrite>. This is the factual chain of events you must keep.

2. Read the writing style in <style\_reference>. Note the tone, verbosity, and language structure. They are sometimes long and so therefore cutoff at 500 characters.

3. REWRITE the content of <logic\_to\_rewrite> using the voice/persona found in <style\_reference>.

4. CRITICAL: 
   
   - You must write in ENGLISH.
   
   - Don't copy the text from <style\_reference> directly; only copy its *style*.
   
   - You must reach the same final answer as the <logic\_to\_rewrite>.
   
   - Please respond in the same JSON format with question, thinking, and model\_response keys.
   
   - The model\_response should be in the same language as the question, irrespective of the <style\_reference>.

5. If you think the initial question and/or answer are flawed, please respond simply "BAD" (no JSON).

</instruction>

\textbackslash{n}

<logic\_to\_rewrite>

\{gemini\_thinking\_trace\}

</logic\_to\_rewrite>

\textbackslash{n}

<style\_reference>

\{style\_reference\}

</style\_reference>
\end{quote}

\section{Example SFT Training Sample}

The example of a synthetic reasoning trace, written in the typical thinking style of Qwen3, which often starts with ``Okay, the user is asking ...'', identifies the language of the question, and uses ``Hmmm''.

\begin{quote}

Okay, the user is asking 'Melyik évben adták ki a 'Dzeelda' which is Hungarian for 'In which year was 'Dzeelda' released?'. Hmm, 'Dzeelda' doesn't sound quite right—I'm thinking this might be a phonetic misspelling. The iconic Nintendo franchise is actually 'The Legend of Zelda', and in Hungarian contexts, it's commonly shortened to 'Zelda' but sometimes gets mangled in pronunciation. Let me verify this.

First, I should confirm which specific game they're referencing since 'Zelda' alone could mean multiple titles. Given how the question is phrased, they're likely asking about the most culturally significant entry—Ocarina of Time, {...}

\end{quote}

Note that this sample is not necessarily a cross-script question, it's meant simply to nudge the LLM to reason more carefully.

\newpage 

\section{LoRA SFT Details} \label{loradetails}

We detail here the configurations for the LoRA SFT with Llama-Factory. For all models, we use the adafactor optimizer and cosine LR scheduling with warmup. In the table below, ``All Attn'' signifies that LoRA modules were applied to the $Q_{proj}$, $K_{proj}$, $V_{proj}$, and $O_{proj}$ weights in the attention block, but not to any FFN parameters. At the time of the experiments, Llama-Factory does not have the capability to apply LoRA modules to the Qwen3 MoE implementation of the FFN layers.

\begin{table}[h]
\centering
\caption{Hyperparameters for the LoRA training runs for which results are reported in Table~\ref{sft_results}.}
\label{tab:hyperparameters}
\begin{tabular}{@{}lcc@{}}
\toprule
\textbf{Hyperparam.} & \small{\textbf{\gpt}} & \small{\textbf{\qwenM}} \\
\midrule
Num. Samples            & 599 & 387 \\
Target Modules          & All Attn & All Attn \\
LoRA Rank ($r$)         & 32                & 32               \\
LoRA Dropout            & 0            & 0.05              \\
Learning Rate           & $2 \times 10^{-5}$ & $6 \times 10^{-5}$ \\
Epochs                  & 8                & 8                \\
Warmup Ratio            & 0.2 & 0.2 \\
\bottomrule
\end{tabular}
\end{table}

\end{document}